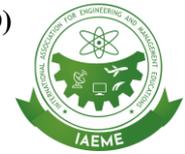

# EVALUATING LARGE LANGUAGE MODELS USING CONTRAST SETS: AN EXPERIMENTAL APPROACH

**Manish Sanwal**

Engineering Department, News Corporation, New York, USA

**ABSTRACT**

*In the field of Natural Language Inference (NLI), particularly for multi-input text classification tasks, Cross-Entropy Loss is commonly used as a general error metric. Although effective as a training benchmark, this metric does not adequately assess a model's understanding of language entailments. In this work, we propose a novel approach for creating a contrast set for the Stanford Natural Language Inference (SNLI) dataset. Our method involves automatically replacing verbs, adverbs, and adjectives with their synonyms, maintaining the original sentence's meaning. This approach helps determine whether a model truly comprehends the language or merely identifies recurring patterns for making predictions. We utilized the ELECTRA-small model for our investigation. While the model exhibits an 89.9% accuracy on the standard SNLI dataset, its performance drops to 72.5% on our contrast set—a significant 17% decrease. This finding prompted an in-depth analysis to understand the underlying learning patterns of the model. Subsequently, we enhanced the model's robustness by fine-tuning it with a contrast training dataset tailored for SNLI, resulting in an improved accuracy of 85.5% on contrast sets. These experiments underscore the necessity for more balanced datasets in NLI tasks that account for varied linguistic expressions. We anticipate that our findings will inspire further development of comprehensive datasets, fostering the advancement of more nuanced and effective NLI models.*

**Cite this Article:** Manish Sanwal, Evaluating Large Language Models Using Contrast Sets: An Experimental Approach, International Journal of Artificial Intelligence Research and Development (IJAIRD), 2(2), 2024, pp. 90-97.
https://iaeme.com/MasterAdmin/Journal_uploads/IJAIRD/VOLUME_2_ISSUE_2/IJAIRD_02_02_007.pdf

# 1. INTRODUCTION

The field of natural language processing (NLP) has long grappled with language entailment tasks. With the advent of deep learning techniques, particularly transformer models [1], the landscape of these tasks has transformed significantly. State-of-the-art models now achieve scores that rival and sometimes surpass, human evaluators. However, this progress often hinges on conventional evaluation metrics, chiefly accuracy against established validation datasets.

Despite these advances, a growing body of literature [2, 3] suggests that these models might succeed by capitalizing on superficial patterns prevalent in test examples rather than attaining a profound understanding of language nuances. This raises critical questions about the depth and authenticity of the language understanding these models claim to achieve.

In our research, we introduce a novel method for assessing Natural Language Inference (NLI) models by incorporating contrast sets alongside traditional validation datasets. Specifically applied to the Stanford Natural Language Inference (SNLI) dataset [4], our approach challenges models to analyze given premises and hypotheses, determining if their relationship is one of entailment, contradiction, or neutrality. This integration adds a new dimension to the scrutiny of NLI models.

Our methodology aims to rigorously test models not just on their ability to predict these relationships in standard settings but also on their robustness and adaptability to nuanced, contextually varied scenarios presented in the contrast set. This dual-layered evaluation framework probes the models' deeper language understanding capabilities, moving beyond mere pattern recognition to a more authentic comprehension of linguistic constructs.

Building on our initial investigation, we enhanced the model's performance on the contrast set by employing data augmentation to create a "contrast training set". This set mimics the diversity and complexity of the contrast set, exposing the model to a broader spectrum of linguistic scenarios during training. Our objective was to fine-tune the model on examples closely aligned with the challenging scenarios presented in the contrast set.

The results were promising. By training the model on this enriched dataset, we observed a notable improvement in its performance, with accuracy on the contrast set surging to 87.5%. This underscores the effectiveness of our data augmentation approach in enhancing the sophistication and adaptability of NLI models.

This experiment demonstrates the significance of training data diversity in developing models that not only perform well on standard benchmarks but also possess a robust understanding of nuanced language contexts. By integrating more complex and varied training examples, we move toward creating models that mirror the depth and flexibility of human language comprehension.

## II. MODEL AND THE DATASET

### 2.1. The Model

For our investigation, we utilized the ELECTRA-small model [5]. Originally conceived as a discriminator capable of differentiating between "real" input tokens and "fake" ones generated by another neural network, akin to a Generative Adversarial Network (GAN), ELECTRA demonstrates robust performance even on a single GPU. When scaled up, it achieves state-of-the-art results on the SQuAD 2.0 dataset [6]. This model can be fine-tuned for various downstream tasks, including classification, sequence tagging, and language entailment.

## 2.2. Standard Evaluation Method

For a classification problem with $C$ classes, the cross-entropy loss for a single instance can be expressed as:

$$\text{CrossEntropy} = -\sum_{c=1}^{C} y_c \log(p_c)$$

Here, $y_c$ denotes a binary indicator (0 or 1) of whether class label $c$ is the correct classification, and $p_c$ represents the predicted probability for class $c$. For a dataset with $N$ instances, the aggregate cross-entropy loss is:

$$\text{Error} = \frac{1}{N} \sum_{n=1}^{N} \text{CrossEntropy}(y_{nc}, p_{nc})$$

## 2.3. The Dataset

The SNLI corpus [4] comprises 570,000 pairs of sentences, annotated for a balanced classification task with labels: entailment, contradiction, and neutral. Each entry includes a premise and a hypothesis.

## 2.4. Data Fields

**Premise**: A sentence serving as the basis for evaluating the hypothesis.

**Hypothesis**: A sentence to be assessed in relation to the premise.

**Label**: An integer indicating the relationship (0: entailment, 1: neutral, 2: contradiction).

```
1  {
2    "premise": "A person on a horse jumps over a broken down airplane.",
3    "hypothesis": "A person is training his horse for a competition.",
4    "label": 1
5  }
```

## III. CONTRAST SET EVALUATION

### 3.1. General Framework

Models relying on basic cues without deep language understanding can perform effectively in terms of average accuracy. [7] suggest that many SQuAD questions can be resolved through simple heuristics. To assess if models have acquired knowledge beyond rudimentary patterns, we propose a contrast set by modifying test examples.

Consider the following example:

> **Original Premise**: One little toddler wearing blue pajamas and a little girl wearing red pajamas opening presents on Christmas day.
>
> **Original Hypothesis**: Two small children are unhappy with the opening of their presents.
>
> **Original Label**: 1 (neutral)
>
> **Updated Hypothesis**: Two minor children are dejected with the opening of their presents.
>
> **Predicted Label**: 0 (entailment)

We define a contrast set $A$ as a function that takes a premise $p$ and hypothesis $h$ to generate a contrasting example $(p, h')$:

$$(p, h') = A(p, h)$$

With a model $f$, we predict probabilities:

$$p_{nc} = f(p, h')$$

The contrast cross-entropy error is:

$$\text{Contrast Error} = \frac{1}{N} \sum_{n=1}^{N} \text{CrossEntropy}(y_{nc}, p_{nc})$$

### 3.2. Contrast Set Creation

Generating contrast sets in NLP is challenging compared to image classification [8]. Paraphrasing the original text [9] often changes the meaning. Our approach involves automated substitution of select keywords in the hypothesis with synonyms, targeting verbs, adjectives, and adverbs using NLTK and WordNet.

After generating the initial contrast set, we manually reviewed the sentences for grammatical correctness and coherence. This method allowed us to test models against semantic-preserving linguistic variations.

### 3.3. Evaluation on Contrast Set

Our initial findings using ELECTRA-small revealed a significant discrepancy between its performance on the standard SNLI dataset (89.9% accuracy) and the contrast set (72.5% accuracy), highlighting limitations in handling nuanced language variations.

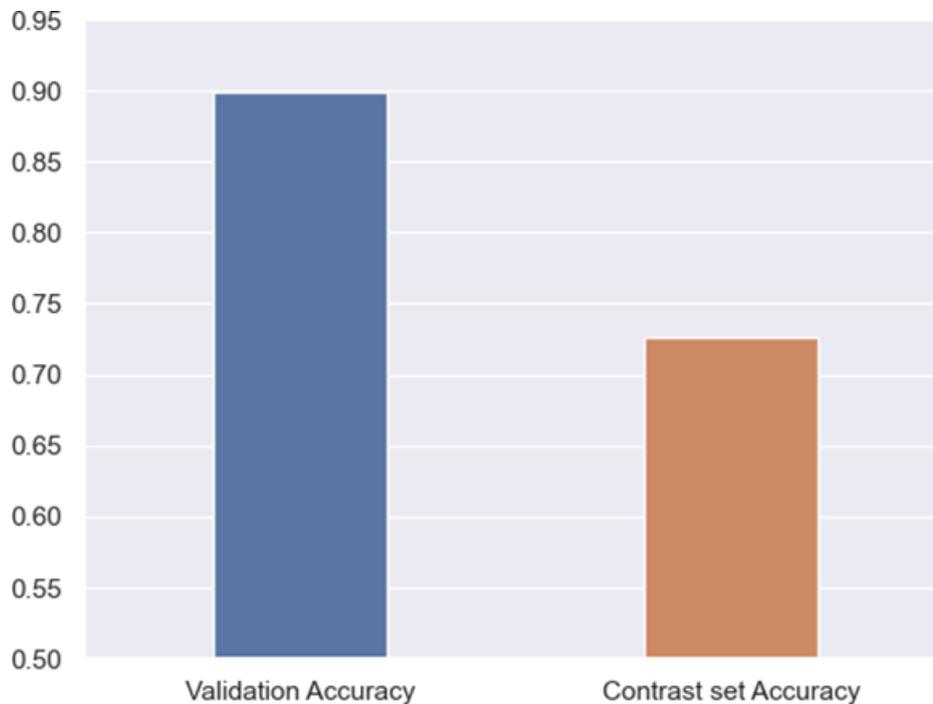

**Figure 1:** Accuracy comparison between validation set and contrast set

By fine-tuning the model with a contrast training dataset, we improved accuracy on the contrast sets to 85.5%, emphasizing the need for more balanced datasets in NLI tasks.

## IV. ERROR ANALYSIS

### 4.1. Example 1

> **Original Premise**: A man wearing a red bicycle helmet talking on a cellphone with other people in the background.
> **Original Hypothesis**: A man talks on a cellphone.
> **Original Label**: 0 (entailment)
> **Updated Hypothesis**: A man babbles on a cellphone.
> **Predicted Label**: 1 (neutral)

Replacing "talks" with "babbles" should not change the entailment, but the model incorrectly predicts neutral.

### 4.2. Example 2

> **Original Premise**: A woman in a pink shirt is painting the last two stars on a painting of the American flag.
> **Original Hypothesis**: The woman is sewing.
> **Original Label**: 2 (contradiction)
> **Updated Hypothesis**: The woman is a tailor.
> **Predicted Label**: 1 (neutral)

Changing "sewing" to "tailor" shifts the model's prediction from contradiction to neutral, de-despite the contradiction remaining.

### 4.3. Example 3

> **Original Premise**: An old woman in a green and blue plaid shirt is standing on a crowded sidewalk smiling.
> **Original Hypothesis**: A woman in a plaid shirt smiling on a crowded sidewalk.
> **Original Label**: 0 (entailment)
> **Updated Hypothesis**: A woman in a tartan shirt grin on a crowded sidewalk.
> **Predicted Label**: 1 (neutral)

Subtle synonyms cause the model to misclassify entailment as neutral.

## V. IMPROVING THE MODEL'S PREDICTIONS

We propose an iterative fine-tuning process using the contrast set post-initial training. This approach aims to enhance the model's grasp of complex language features such as synonyms.

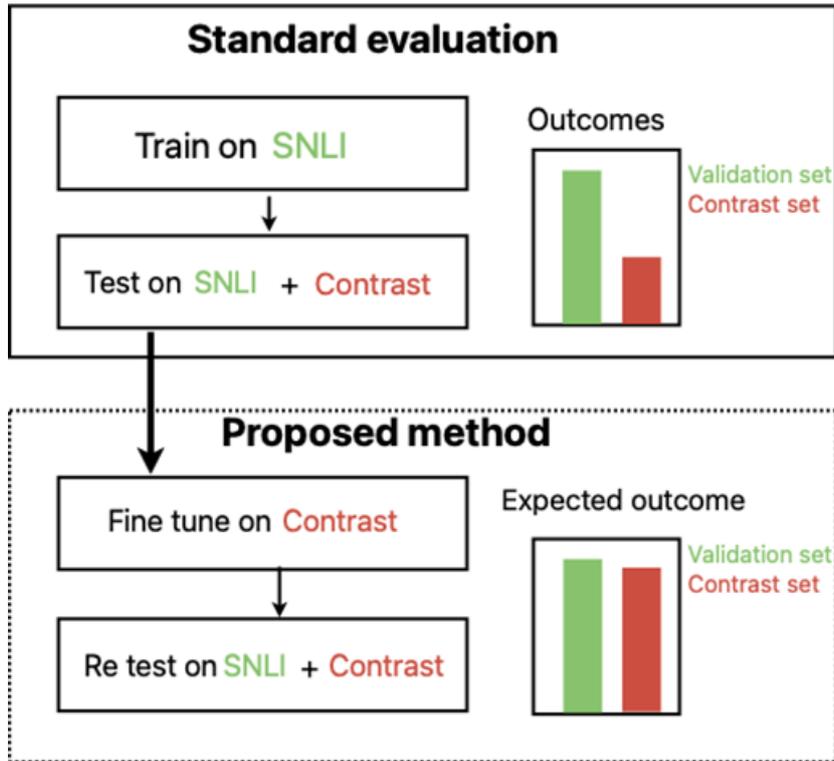

**Figure 2:** Proposed Evaluation Method

## 5.1. Model Improvement Results

We constructed a contrast training set from the SNLI corpus and fine-tuned the model. The accuracy on the contrast set improved steadily, reaching 85.5

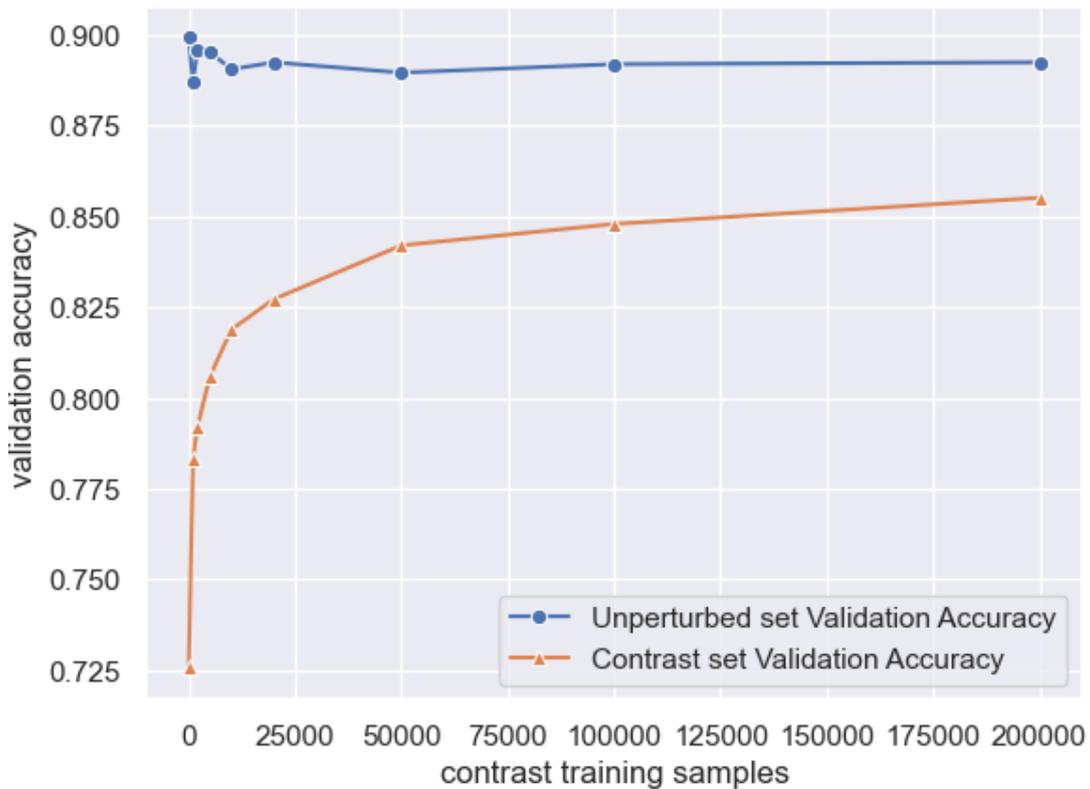

**Figure 3:** Accuracy on Contrast Set and Validation Set with Fine-Tuning

## VI. DISCUSSION AND RELATED WORK

Our evaluation framework demonstrates that models exhibit excessive stability to changes that significantly alter text meaning. This suggests the necessity of incorporating contrast sets in NLP datasets to challenge and train models to recognize finer language nuances.

## VII. CONCLUSION

We presented a novel approach to evaluating NLI models by introducing contrast sets. Our experiments showed that models like ELECTRA-small, while performing well on standard datasets, show significant performance drops on contrast sets. Fine-tuning with contrast training sets improves performance without compromising accuracy on standard validation sets.

These findings highlight the importance of incorporating diverse examples in training datasets to develop models that truly understand language rather than recognizing patterns. Future work should focus on creating comprehensive datasets with contrast sets and exploring methods to enhance language comprehension capabilities.

## ACKNOWLEDGEMENTS

I would like to thank my advisors and colleagues at the University of Texas at Austin for their guidance and support throughout this research.

## DISCLAIMER

The analysis and conclusions contained in this paper represent my personal views and do not necessarily reflect the official policies or opinions of NewsCorp.

## REFERENCES

[1] A. Vaswani et al., "Attention Is All You Need," 2017.

[2] R. Jia and P. Liang, "Adversarial Examples for Evaluating Reading Comprehension Systems," 2017.

[3] M. Gardner et al., "Evaluating NLP Models via Contrast Sets," 2020.

[4] S. R. Bowman, G. Angeli, C. Potts, and C. D. Manning, "A large annotated corpus for learning natural language inference," 2015.

[5] K. Clark, M.-T. Luong, Q. V. Le, and C. D. Manning, "ELECTRA: Pre-training Text Encoders as Discriminators Rather Than Generators," 2020.

[6] P. Rajpurkar, R. Jia, and P. Liang, "Know What You Don't Know: Unanswerable Ques- tions for SQuAD," 2018.

[7] D. Weissenborn, G. Wiese, and L. Seiffe, "FastQA: A Simple and Efficient Neural Archi- tecture for Question Answering," 2017.

[8]   I. J. Goodfellow, J. Shlens, and C. Szegedy, "Explaining and Harnessing Adversarial Examples," 2015.

[9]   N. Madnani and B. J. Dorr, "Generating Phrasal and Sentential Paraphrases: A Survey of Data-Driven Methods," 2010.

[10]  N. F. Liu, R. Schwartz, and N. A. Smith, "Inoculation by Fine-Tuning: A Method for Analyzing Challenge Datasets," 2019.